\documentclass[11pt]{article}
\pdfoutput=1
\usepackage[margin=1.2in]{geometry}
\usepackage{graphicx}

\usepackage[T1]{fontenc}

\usepackage{verbatimbox}
\usepackage{hyperref}
\hypersetup{
    colorlinks=true,
    linkcolor=blue,
    filecolor=magenta,      
    urlcolor=blue,
}

\usepackage{gensymb}
\usepackage{booktabs}
\usepackage{adjustbox}
\usepackage{lscape}
\usepackage{subcaption}
\usepackage{caption}
\usepackage{multirow}
\usepackage{hyperref}
\usepackage[nodisplayskipstretch]{setspace}
\newtheorem{definition}{Definition}
\usepackage{algorithm}
\usepackage{algpseudocode}
\usepackage{booktabs}
\usepackage[colorinlistoftodos]{todonotes}
\usepackage{amssymb}
\usepackage[utf8]{inputenc}
\usepackage{color}
\usepackage{amsmath}

\usepackage{algorithm}
\usepackage{algpseudocode}
\usepackage{txfonts}

\usepackage{natbib}

\graphicspath{ {Pictures/} }

\setcitestyle{authoryear,open={(},close={)}}
 
\newcommand{\ignore}[1]{}

\usepackage{setspace}
\usepackage{gensymb}
\usepackage[margin=1cm]{caption}

\begin{document}



\title{Adjusting Rate of Spread Factors through Derivative-Free Optimization: A New Methodology to Improve the Performance of Forest Fire Simulators}

\author{Jaime Carrasco$^{a}$, Cristobal Pais$^{b}$ \\ Zuo-Jun Max Shen$^{b}$, Andr\'es Weintraub$^{a}$}

\maketitle  

\noindent
{$^{a}$University of Chile, Industrial Engineering Department\\ 
 $^{b}$University of California Berkeley, IEOR Department}

\begin{abstract}
\noindent
In practical applications, it is common that wildfire simulators do not correctly predict the evolution of the fire scar. Usually, this is caused due to multiple factors including inaccuracy in the input data such as land cover classification, moisture, improperly represented local winds, cumulative errors in the fire growth simulation model, high level of discontinuity/heterogeneity within the landscape, among many others. Therefore in practice, it is necessary to adjust the propagation of the fire to obtain better results, either to support suppression activities or to improve the performance of the simulator considering new default parameters for future events, best representing the current fire spread growth phenomenon. In this article, we address this problem through a new methodology using Derivative-Free Optimization (DFO) algorithms for adjusting the Rate of Spread (ROS) factors in a fire simulation growth model called Cell2Fire. To achieve this, we solve an error minimization optimization problem that captures the difference between the simulated and observed fire, which involves the evaluation of the simulator output in each iteration as part of a DFO framework, allowing us to find the best possible factors for each fuel present on the landscape. Numerical results for different objective functions are shown and discussed, including a performance comparison of alternative DFO algorithms.
\end{abstract}



\section{Introduction}
 
Considerable efforts have been made in recent decades to simulate fire through a heterogeneous forest landscape due to the increase of such events as a result of global warming and human carelessness \citep{Running2006,Westerling2006, Westerling2016}. To date, a wide set of fire growth simulators are available, which utilize a range of different modeling approaches and underlying fire behavior prediction systems to simulate the fire spread dynamics based on demographic, topographic, and environmental conditions. An excellent review of these can be found in \cite{SimReview}.

One of these computational tools is Prometheus, a deterministic fire growth simulator released in May 2009 \citep{Prometheus}. Using spatial fire behavior input data on topography (slope, aspect, and elevation) and the Canadian Forest Fire Behavior Prediction (FBP) System fuel types along with an hourly weather stream, it simulates fire growth based on the Huygens’ principle of wave propagation. Another simulator with similar characteristics is FARSITE \citep{Finney2005} based on the American BEHAVE System. As indicated in \cite{SimReview}, these two models obtain the best simulations of historical fires: FARSITE in the United States and Prometheus in Canada. 

Although Prometheus and FARSITE have excellent performance in general, they are not suitable for fire-smart forest management that requires that both the harvest and simulation models have a well-structured and natural interface for easily exchanging data at each iteration of the process, in order to incorporate decision-making modules. This becomes even more relevant in a multistage framework in which multiple events, whether harvest decisions and/or fires may occur each year for a given planning horizon. For this reason, a new cellular-based fire growth simulator has been recently developed called Cell2Fire \citep{Pais2019}. Developed in C\textbf{++}, it was designed to run both on daily-user machines and High-Performance Computer (HPC) systems. In its current version, it incorporates fuel and fire models from the Canadian Forest Fire Behavior Prediction (FPB) System, allowing the user to simulate the fire dynamics across a grid that represents a real forest landscape.

However, the performance of any simulator and/or computer programs depends to a large extent on its input parameters, calculated as a function of the data provided by the user or following theoretical bases that support their values. This way, some performance problems may be a result of inaccurate data on fuel moistures, fuel descriptions, weather, or improperly represented local winds, among other sources of errors \citep{finney1998farsite}. As is indicated in \cite{Pais2019}, the ability to build a realistic fire scar with Cell2Fire depends on the Rate of Spread (ROS), which causes the fire to advance faster or slower through the cells of the forest. A similar approach is followed by simulators such as FARSITE and Prometheus \citep{finney1998farsite, Prometheus}. Modeling the landscape as a continuous surface, they base their propagation model through the forest on the assumption that under homogeneous conditions, the burned area has an elliptical shape whose deformation mainly depends on the wind speed and its axes are constructed from the main rate of spread values: Head ROS (HROS), Back ROS (BROS) and Flank ROS (FROS). In the case of FARSITE, these parameters are provided by BEHAVE, and obtained from the Canadian FBP System for Prometheus and Cell2Fire.

The noise of the data and different approximation inaccuracies directly affect the propagation model by perturbing the calculation of the different ROS values mentioned above, and thus, this can lead to an over/underestimation of the ROS magnitude, in disagreement with the observed fire.
For these reasons, systems like FARSITE, BEHAVE, WildFire Analyst, and others \citep{rothermel1983field,finney1998farsite,ramirez2011new}, have introduced a module of adjustment factors directly on the ROS, allowing the user to use expert judgment or local data to tune the simulation parameters to observed or actual fire spread patterns. These factors are fuel model specific, multiplied by the rate of spread to achieve the specified adjustment \citep{finney1998farsite,ramirez2011new}. The latter is done manually in FARSITE, without any guarantee that the new factors configuration will improve the prediction. Moreover, the complexity of searching for factors increases along with the heterogeneity of the forest, directly depending on the number of fuels inside it. A recent paper \citep{srivas2017data} addresses this problem in FARSITE, introducing an automatic calculation using an Ensemble Kalman Filter that exploits the uncertainty information on the simulated fire perimeter, fuel adjustment factors, and measured fire perimeters. On the other hand, \citep{cardil2019adjusting} proposes a method to determine the optimal adjustment factors by fuel model from the fire observed in real-time
in order to minimize the arrival time error with respect to the simulated fire.

The main objective of this article is to show how a novel methodology based on Derivative-Free Optimization can serve to improve the performance of fire simulators in both real-time and future events simulations through an automatic adjustment of the Rate of Spread factors, obtaining more accurate fire scars. For this, we use Prometheus simulator as a proxy because it provides us the evolution of fire scars hour-by-hour with the purpose of imitating the behavior of a fire in real-time. Using this information, we minimize an objective function that represents the error in the fire spread evolution, having the ROS factors as the decision variables of the optimization model. Also, in order to show the adjustment approach for future events, we analyze a case study in Canada where a landscape located in the Alberta Region suffered a large fire on September 25, 2001.

The paper is organized as follows: In section 2, the main fire spread simulation logic of Cell2Fire is described and a brief description of Derivative-Free Optimization algorithms is given. Section 3 introduces the Derivative-Free optimization framework and the methodology approach to adjust the ROS factors in Cell2Fire, analyzing the fire spread and growth dynamics of each fuel type included in the FBP system, as well as describing the main algorithms implemented. In section 4, optimization results are discussed for a case study based on a real forest in Canada and a comparison of the performance with different sets of parameters is included for several (non)-homogeneous test instances. Finally, section 5 contains the conclusions and future work of the project. 


\section{Methods} 
\label{S:2}
\subsection{Cell2Fire Simulator}
Cell2Fire is a fire growth cell-based simulator developed in C++. It allows the user to simulate the fire dynamics inside a grid instance that can represent a real forest based on variables such as fuel type of each cell, weather, elevation (topographic/terrain components), ignition points, and all the aspects considered by the FBP System -- see more detail in \citep{Pais2019}. 

In algorithmic terms, Cell2Fire simulates the growth of fire by tracking the state of all cells as the model progresses through discrete time steps on a raster grid where a cell can be in one state \textit{``Available''}, \textit{``Burning''}, \textit{``Burned''}, \textit{``Harvested''}, or \textit{``Non-Fuel''}. Fires begin with randomly occurring lightning strikes that will ignite a cell or an ignition point can be chosen deterministically to reproduce, for example, a historical event. At each time step, the fire may spread to neighboring cells based on calculations of the Head Rate of Spread (HROS), Flank Rate of Spread (FROS, identical to both sides) and Back Rate of Spread (BROS) obtained from the Canadian FBP System. Subsequently, a messaging signal process is initiated between the neighboring cells. In the current implementation, each cell has (at most) 8 neighbors. Fire progresses through each available axis of the current burning cell towards the center of the adjacent cells, starting a new fire when it reaches another cell's center. Thus, a higher rate of spread (ROS) entails that there is a greater chance of fire spreading to a neighboring cell. 

\begin{figure}[h!]
	\centering
    \includegraphics[scale=1.00]{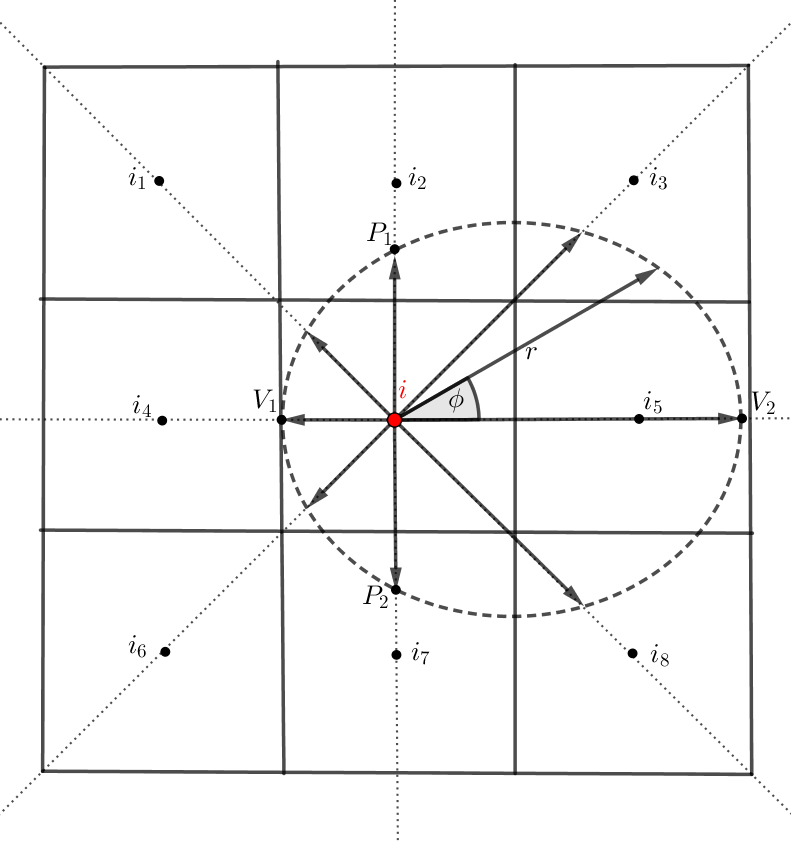}
    \caption{Elliptic approach for $ WD = 0 \degree $. ROS ($r$) is calculated for each angle $\phi$.}
    \label{Ellipse}
\end{figure}

As mentioned in \cite{Pais2019}, Cell2Fire is based on an adaptation of the elliptical model proposed in \cite{ORegan1973,ORegan1976} using the main Rate of Spread components of the FPB System (HROS, FROS, and BROS). It considers the center of a cell as a focus of an ellipse and an adjustment of the ROS to the generated ellipse is performed as shown in Figure \ref{Ellipse}. The implemented ROS distribution scheme is as follows:

\begin{itemize}
\item[i.] Parameterize the ellipse from one of its foci, using polar coordinates $\left(r,\phi\right)$ where $r$ is the length of the radius vector and $\phi$ represents its angle with respect to the horizontal line (0\degree East).

\item[ii.] Set the main parameters of the ellipse (axes and eccentricity) at time $t$:
    \begin{eqnarray}
        a &=& \dfrac{HROS + BROS}{2} \times t,\\
        b &=& \dfrac{2 \times FROS}{2} \times t,
    \end{eqnarray}
    where  $FROS = \dfrac{HROS + BROS}{2 LB}$ and
    $LB = \dfrac{a}{b}$, the length-to-breadth ratio. Therefore, the eccentricity is calculated by: 
    \begin{eqnarray}
        e = \sqrt{ 1- \left( \dfrac{{(FROS \times t)}^{2}}{{ {\dfrac{(HROS + BROS) \times t}{2}}}^{2}} \right)   }
    \end{eqnarray}
    to expand the ellipse generated by the fire propagation at time $t$ on the two main axes.

\item[iii.] Calculate the Rate of Spread $r$ as a function of the angle $\phi$:
$$r=\dfrac{a\left(1-e^{2}\right)}{\left(1-e\cdot cos\phi\right)}. $$

\end{itemize}

Therefore, the full dynamic of the fire is determined by these three magnitudes: HROS, BROS, and FROS plus a fourth related to the eccentricity of the ellipse. Despite its practical usage, this elliptical model may have its limitations for practical situations as pointed out by \cite{richards1993properties}: local wind fluctuations may decrease the $LB$ value and then overestimate the heading spread of a fire at the expense of flanking spread. For this reason, \cite{finney1998farsite} considers pertinent to introduce some type of compensation through the use of spread rate adjustment factors. The above justifies the introduction of four ROS adjustment factors: $x_1, x_2, x_3, x_4$. These variables internally multiply the rate of spreads: $HROS$ ($x_1$), $FROS$ ($x_2$), and $BROS$ ($x_3$) obtained by Canadian FBP module; and a additional factor $ECC$ ($x_4$) respective to adjust the ellipse eccentricity.

\subsection{Derivative-Free Optimization}

Derivative-free optimization (DFO) is an area of nonlinear optimization that deals with problems where the derivatives of the objective function (and potentially, constraints) are not available.
Due to a growing number of applications in science and engineering, the development of DFO algorithms has increased and given greater attention in recent decades. Some applications using DFO algorithms can be found in \citep{alarie2013snow,alexandridis2008cellular,begin2010dfo,hare2010using}.

\begin{figure}[h!]
	\centering
    \includegraphics[scale=0.52]{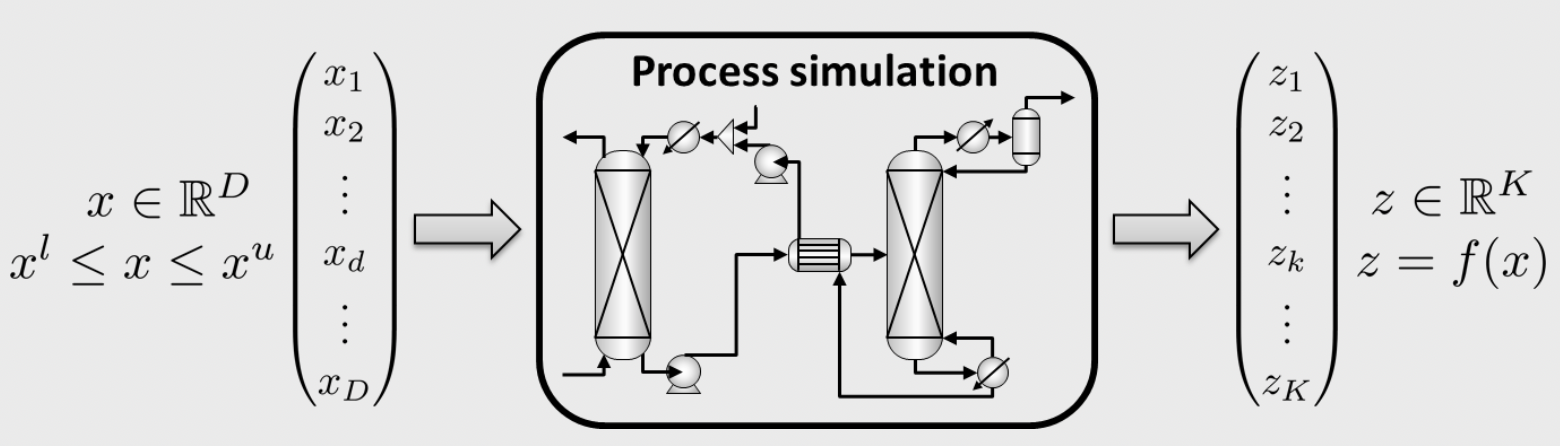}
    \caption{A ``Black-Box System'' assumes that the black box can be queried through a simulation or experimental measurements that provide a system output for specific system input values $x \in \mathbb{R}^D$. A principal challenge in practical optimization is how to optimize an objective function $z=f(x)$ that depends on this process in the absence of an algebraic model.} 
    \label{BBO}
\end{figure}

There are different situations where this methodology is appropriate: i) the functions defining the problem are provided through a computer simulation that cannot be easily subjected to automatic differentiation (see Figure \ref{BBO}); ii) the optimization problem involves conducting a laboratory experiment, with no explicit mathematical expressions; iii) the objective function is noisy and the gradient estimation may be completely useless; iv) when the evaluation of the functions require a significant amount of computational power, it may be prohibitive to perform the necessary number of function evaluations  -- normally no less than the number of variables plus one -- to provide a single gradient estimation.

The diversity of applications includes problems in engineering, mathematics, physics, chemistry, economics, finance, medicine, transportation, computer science, business, and operations research (see e.g. \cite{conn2009introduction,audet2017derivative}). Some examples of them are: Tuning of algorithmic parameters \citep{audet2006finding,begin2010dfo}; Engineering design \citep{booker1998managing, booker1998optimization}; Molecular geometry \citep{alberto2004pattern}; Medical image registration \citep{oeuvray2005trust}; and dynamic pricing \citep{levina2009dynamic}.  

In our research, we will follow the ideas of ``Parameter Fit'' presented in \cite{audet2006finding} and \cite{alexandridis2008cellular}. As we pointed out in the Introduction, most numerical codes (for simulation, optimization, estimation, etc) depend on a number of internal parameters. Researchers implementing numerical algorithms know how critical the choices of these parameters are and how much they influence the performance of solvers. Typically, these parameters are set to values that either have some mathematical justification or satisfactory empirical results. One way to automate the choice of the parameters  --- in order to find possibly optimal values --- is to consider an optimization problem whose variables are the parameters and whose objective function measures the performance of the solver for a given set of parameters, measured by CPU time or by some other indicator such as the number of iterations taken by the solver (see \cite{conn2009introduction}). However, in our study, we are not interested in the CPU time or the number of iterations that Cell2Fire makes to get a more accurate fire scar. Since Cell2Fire simulated scars depend dynamically -- in simulation time -- on the fuel type (ROS obtained from the FBP System), our parameters to adjust/re-scale the fire spread model will be such that they change the magnitude of the ROS among the main propagation axes. This way, our main performance measurement will be the adjustment error with respect to a real/historical fire scar observed, a real-time fire scar provided to predict the most likely evolution of an on-going fire, or a scar simulated by an already calibrated software. 

\begin{figure}[h!]
	\centering
    \includegraphics[scale=0.61]{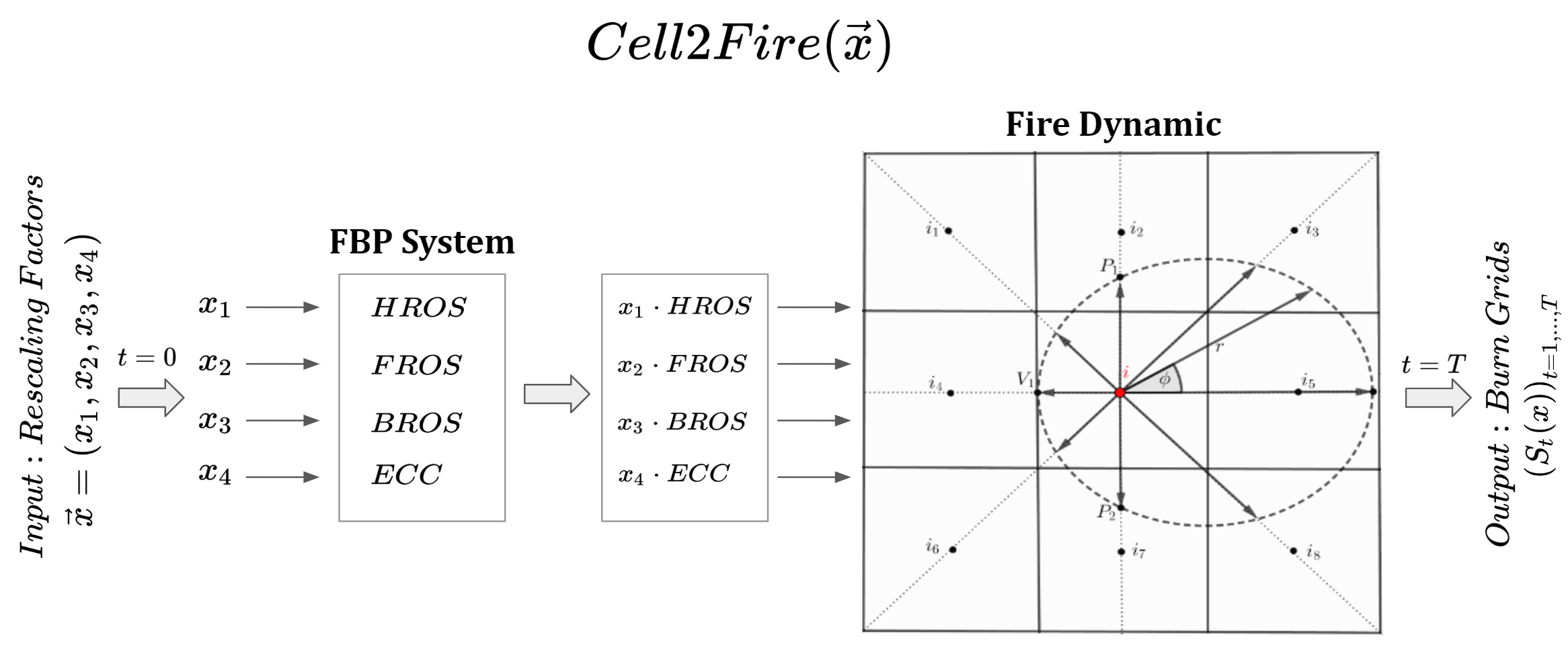}
    \caption{$Cell2Fire(\vec{x})$ Function. $S_t(\vec{x})$ is defined as a function that depends on the variables $x_1$, $x_2$, $x_3$ which modify the magnitude of the parameters HROS, FROS, and BROS obtained by the FBP System, and a fourth variable $x_4$ that adjusts the eccentricity of the ellipse, called ECC factor. Its outputs $S_1(\vec{x})$, $S_2(\vec{x})$,...,$S_T(\vec{x})$ are simulated fire scars for $t=1, t=2,...,t=T$ hours using the parameters $\vec{x} \in \mathbb{R}^4$.}
    \label{Cell2Fire_Function}
\end{figure}

In Figure \ref{Cell2Fire_Function}, we show a diagram of the internal dynamics of Cell2Fire as a function of the ROS factors $\vec{x} = (x_1,x_2,x_3,x_4)$. Note that for $\vec{x} = (1,1,1,1)$, Cell2Fire behaves normally (by default). Let $Cell2Fire(\vec{x})$ be the ``simulator function''. Since we want to compare simulated and observed fire scars for different ROS factors, the $Cell2Fire(\vec{x})$ function maps the $\vec{x}$ vector to burn-grids $S_t(\vec{x})$, for all time steps $t=1,...,T$, where $T$ is the total simulation time and $S_t(\vec{x})$ a 0-1 matrix with the same dimensions as the landscape where components $s^{t}_{(i,j)} = 1$ when a cell -- located at coordinates $(i,j)$ -- was burned before time $t$ and 0 otherwise (see Figure \ref{DFO_Framework}). Thus, $\sum_{i,j}s^{t_1}_{i,j}(\vec{x}) \leq \sum_{i,j}s^{t_2}_{i,j}(\vec{x})$ if $t_1 \leq t_2$. This function represents the core of the DFO framework and will be explained in detail in the next section.

Finally, we mention that, since the magnitude of the ROS depends on the characteristics of the type of forest fuel being burned at time $t$ \citep{Rothermel1972,VanWagner1987}, the adjustment factors may have a dependence by type of fuel \citep{rothermel1983field,finney1998farsite}. In order to address this problem, we naturally extend the initial framework by introducing a vector $\vec{x}_{\omega} = (x_{\omega 1},x_{\omega 2},x_{\omega 3} ,x_{\omega 4} )$ for each ${\omega \in \Omega}$, where $\Omega$ denotes the set of all fuel types existing within the landscape (see Section \ref{FMSapproach} for more details).

\section{Rate of Spread adjustment factors}
\label{S:3}
In this section, we describe the theoretical background of the DFO algorithms, their practical advantages/limitations, as well as how we can exploit their characteristics for modeling our fitting problem as a simple error-minimization problem. Theoretical and computational implementation details are discussed for the selected fitting strategies. 

Based on the discussion from Section \ref{S:2}, the main tunable parameters present in Cell2Fire consist of the adjustment factors for the main ROS components (Head, Back, and Flank ROS) that define the shape and magnitude of the fitted-ellipses, associated with each burning cell. As mentioned in \cite{Pais2019}, this step is critical to reproduce valid and realistic fire scars as well as represent the fire spread dynamic evolution with minimum error. Inappropriate ROS factors values can lead to poor results in both the evolution and final fire scar obtained, specially depending on the structure and characteristics of the forest: (1) different fuel types obtain different accuracy performance --- associated with different ROS estimations, (2) heterogeneous landscapes tend to increase the structural differences between the wave-propagation and the cellular-automata models, and (3) extreme weather conditions significantly affect the behavior of the propagation dynamic due to the differences in the conditions under which fire models algorithms were developed \citep{duff2018conditional}.

Two main optimization approaches are tested in order to find the best ROS adjustment factors $\vec{x}$ and improve the performance of Cell2Fire with respect to an observed fire through the formulation of an optimization problem that minimizes the fire scar evolution error over time. The first one seeks to find four general factors, which we will denote by $\vec{x}=(x_1,x_2,x_3,x_4) \in \mathbb{R}^4_+$, independent of the type of fuel inside the landscape. Thus, each factor is an ``average factor'' that re-scales all ROS independent of the fuel type in order to improve the estimate of the observed scar. This method will be faster in computational terms, but less precise than a fuel-dependent model. The extension is covered in our second model, where we try to address the problem of factors adjustment empirically proposed in \cite{finney1998farsite}, however, we go a little further: we seek not only to adjust the corresponding factor to the HROS, but we also introduce factors for the BROS, FROS, and for the eccentricity of the elliptical model. This way, this is a fuel model specific approach where the factors depend on the type of fuel. Let $\Omega$ be the set of all fuel types, then $\vec{x}_{\omega} = (x_{1\omega},...,x_{4\omega})$ denotes the ROS factors associated with a specific fuel type ${\omega \in \Omega}$. Therefore, the new decision vector in this model is $\vec{x} = (\vec{x}_{\omega})_{\omega \in \Omega} \in \mathbb{R}^{4\left|\Omega\right|}_+$.

\subsection{Global approach}
\label{globalaproach}
\begin{figure}[h!]
	\centering
    \includegraphics[scale=0.65]{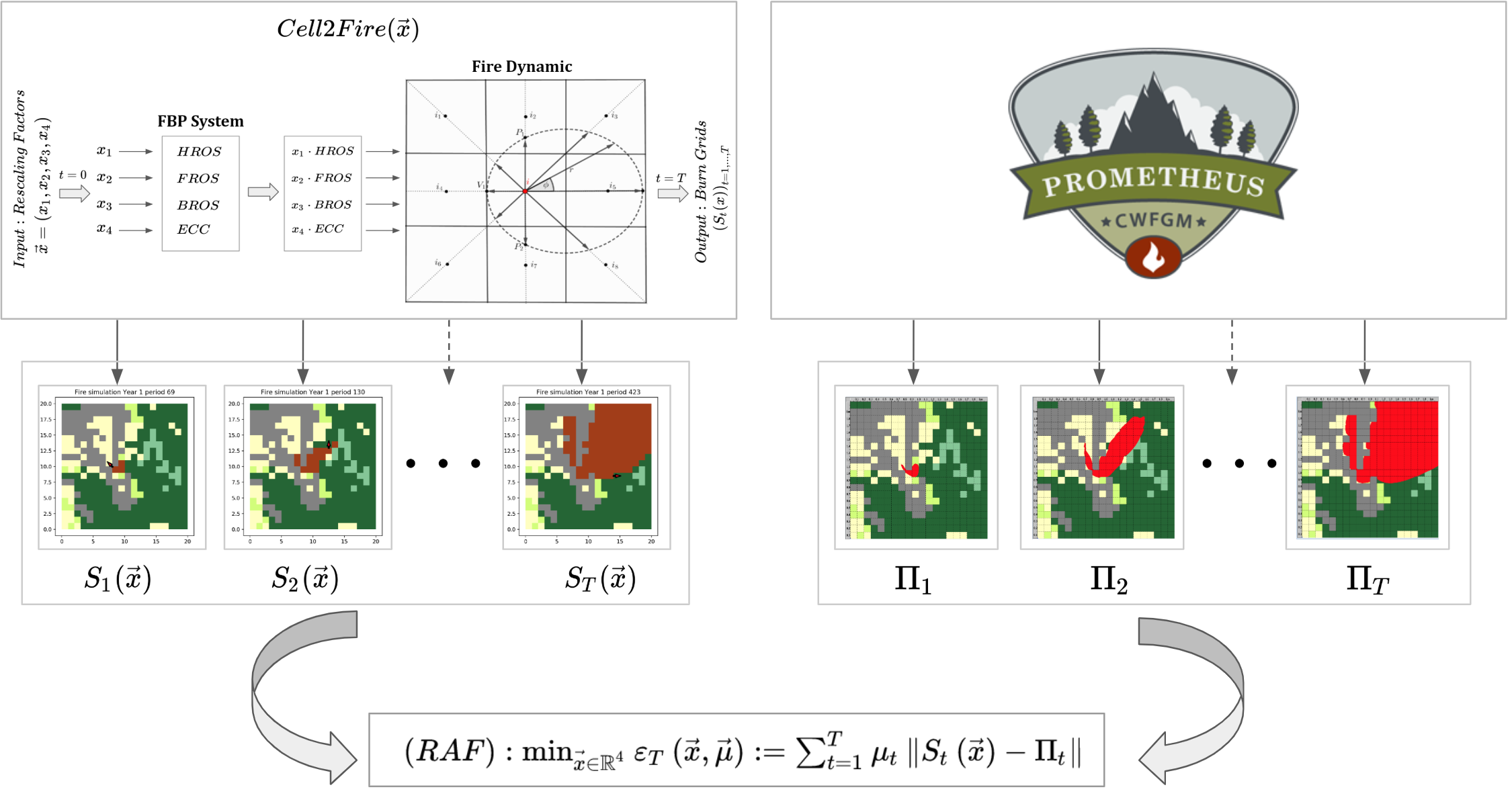}
    \caption{Derivative-Free Optimization framework: This figure represents the scheme of how the (RAF) problem is built. On the left side, Cell2Fire generates the fire scars $S_1(\vec{x})$,..., $S_T(\vec{x})$ which are dependent of the ROS adjustment factor vector $\vec{x}$ and time $t \in T$. On the right side, the observed/simulated fire scars $\Pi_{1}$,..., $\Pi_{T}$ are provided. From this, the Black-Box function $\varepsilon_{T}(\vec{x}, \vec{\mu})$ is formulated as the weighted sum of the errors made using parameters $\vec{x}$ at each time $t$. Finally, the (RAF) problem aims to find those parameters that minimize the total error.}
    \label{DFO_Framework}
\end{figure}

The simplest tuning approach consists of finding the $\vec{x}^*$ vector that minimizes the global fire scar evolution error for a heterogeneous landscape. In this case, a unique $\vec{x}(\vec{\mu}) \in \mathbb{R}^4_+$ vector containing the relevant re-scaling parameters is optimized by solving the ROS Adjustment Factor \textit{(RAF)} problem, using a set of desired weights $\vec{\mu} \in \mathbb{R}^{|T|}_+$ depending on the relevance of each period $t$ accuracy for the research purpose: e.g. get the best accuracy during the first hours after the fire ignition or obtain the most likely final fire scar (see Figure \ref{DFO_Framework} and Definition \ref{Def:1}). This is equivalent to equally modify the shape and magnitude of the fire ellipses generated by each burning cell inside Cell2Fire.


Based on the previous discussion, we introduce the following notation and definitions:
\begin{itemize}
\item[-] Let the subscript $t \in T$ be the simulation time measured in hours and $T$ the discrete set containing the hours to simulate (simulation horizon).  Our methodology allows other time intervals depending on the evolution of the observed fire. For simplicity, we use one hour in this study as the main time-step unit.

\item [-] A vector $x \in \mathbb{R}^4$ is denoted by $\vec{x}$ or $\pmb{x}$, representing the four ROS adjustment factors.

\item[-] Let $S_{t}:\mathbb{R}^{4}\rightarrow\left\{ 0,1\right\} ^{m\times n}$ be a function which outputs are the burn-grids obtained from the $Cell2Fire(\vec{x})$ ``simulator function'', dependent on the parameters $\vec{x}=\left(x_{1},x_{2},x_{3},x_{4}\right)$, where $x_{1}$, $x_{2}$, $x_{3}$, and $x_{4}$ are the re-scaling factors that multiply the main rate of spread values for the ellipse-fitting procedure of Cell2Fire: \textit{HROS}, \textit{FROS}, and \textit{BROS}, and an additional parameter for $ECC$ factor, respectively (see section \ref{Cell2Fire_Function}). 

\item[-] The $S_t(\vec{x})$ function maps the four main ROS factors into a series of Burn-Grids, each one associated with a certain period $t$ containing the simulated fire scar up to that hour. For a certain period $t$ and a forest containing $m \times n $ cells, each Burn-Grid consists of a $m \times n$ binary matrix where entries containing $1s$ represent burned cells while the ones with $0s$ represent available cells. Therefore, the output of $S_{t}\left(1,1,1,1\right)$ is equal to the binary matrix obtained without any adjustment on the ROS values, i.e. using the direct output from the FBP system.

\item[-] Let $\Pi_{t}$ be the Burn-Grid binary matrix generated by Prometheus after $t$ hours of simulation or a historical wildfire scar. Similar to $S_t(\vec{x})$, this matrix represents the burned cells (with value 1) up to time $t$.

\item[-] Objective Function: we seek to minimize the hourly fire scar evolution error, ideally using observed fire data (when available). In our study, we use Prometheus' outputs as a proxy. 
\end{itemize}

Using the above notations and observations, we define the following optimization problem which seeks to find the optimal ROS adjustment factors $\vec{x}^{*}=\left(x_{1}^{*},x_{2}^{*},x_{3}^{*}, x_{4}^{*}\right)$.

\begin{definition}
\label{Def:1}
Let $\varepsilon_{T}(\vec{x}, \vec{\mu})$ be the black-box error function for a simulation horizon $T$, $S_t(\vec{x})$ the simulator function at time-step $t$ and a ROS adjustment factor vector $\vec{x}$, $\Pi_{t}$ the expected binary matrix at period $t$, $||.||$ a matrix norm function (e.g. Frobenius norm), and $\vec{\mu} = (\mu_1, ..., \mu_{T})$ the vector of weights associated with each time-step.
We define the \textbf{ROS adjustment factor (RAF) problem} for fitting the values of the $\vec{x}$ vector as:

\[
\left(RAF\right):\min_{\vec{x}\in\mathbb{R}^{4}}\varepsilon_{T}\left(\vec{x}, \vec{\mu}\right):=\sum_{t=1}^{T}\mu_{t}\left\Vert S_{t}\left(\vec{x}\right)-\varPi_{t}\right\Vert .
\]
\end{definition}

\smallskip{}
\newpage

\textbf{Observations:}
\begin{itemize}
	\item[-] The term $\left\Vert S_{t}\left(\vec{x}\right)-\Pi_{t}\right\Vert $ measures the error between Cell2Fire and the provided fire scars at time $t$ hours using the $\vec{x}$ ROS parameters. Important is to emphasize at this point that in case of having the evolution of the observed fire scar, it would not be necessary to use simulated ones (e.g. from Prometheus), which would improve the accuracy of the adjustment for practical cases.
    
	\item[-] $\varepsilon_{T}: \mathbb{R}^{4}\rightarrow\mathbb{R}_{+}$ is a function that measures the error between the simulated and observed fire scar evolution. In order to obtain it, Cell2Fire must be executed for the corresponding parameters $\vec{x} \in \mathbb{R}^4$.
	
	\item[-] Different weight vectors $\vec{\mu}$ can be selected for different fitting objectives, e.g. minimize global hourly error or give specific emphasis on final scar differences, adding flexibility for analyzing different types of decisions -- operational, tactical, and strategical.
	
	\item[-] The \textit{RAF} problem is an optimization problem that cannot be solved with conventional methods because we do not have any information about the derivatives or the algebraic structure of the function $\varepsilon_{T}\left(\vec{x}, \vec{\mu}\right)$.
\end{itemize}

Despite the explicit computational advantage of this scheme with respect to the individual optimization of each fuel type factor, one of the main limitations of this approach consists of the fact that the optimal $\vec{x}^{*}$ vector is instance-dependent and thus, it is not likely useful for different forest compositions and fuel type distributions.

\subsection{Fuel Model Specific (FMS) Approach }
\label{FMSapproach}

In this section, we denote the ROS adjustment factors vector by  $\vec{x} = (x_{1\omega},...,x_{4\omega})_{\omega \in \Omega}$  where $\Omega$ denote the set of all fuel types, i.e. we suppose that  the number of variables in the \textit{RAF} problem increase to $4\cdot\left|\Omega\right|$. We proceed to find the optimal vector $\vec{x}^{*}$ such that the combined fuel types dynamic is adjusted by these parameters, minimizing the error in the fire evolution of a specific heterogeneous landscape instance. Then, $\vec{x}^{*}$ is the solution of 
\[
\left(RAF_{FMS}\right):\min_{\vec{x}\in\mathbb{R}^{4\left|\Omega\right|}_+}\varepsilon_{T}\left(\vec{x}, \vec{\mu}\right):=\sum_{t=1}^{T}\mu_{t}\left\Vert S_{t}\left(\vec{x}\right)-\varPi_{t}\right\Vert .
\]

Thanks to this approach, we obtain the most accurate parameters for each fuel type, minimizing the simulation error of the internal fire spread model.

This approach requires more computational resources than the global approach presented in section \ref{globalaproach} in order to obtain the optimal $\vec{x}^{*}(\pmb{\mu})$ vectors since we are increasing the number of adjustable parameters by a factor of $|\Omega|-1$. However, it arises as one of the most precise approaches for very heterogeneous forest during our experiments, as we will discuss when applying it to our case study (see Section \ref{S:4}).



\subsection{DFO algorithms}
\label{Algos}
In order to solve the (RAF) problem, we apply a series of powerful and easy-to-implement DFO algorithms following the techniques and recommendations from \cite{conn2009introduction}  and \cite{audet2017derivative}. Based on the characteristics of our problem and the expected performance of the different algorithms (convergence to the global optimum is not guaranteed), we implement, test, and compare the following algorithms in order to find the optimal $\vec{x}$ parameters:

\begin{itemize}
  
  \item[-] \textbf{Nelder-Mead}: an algorithm introduced in \cite{nelder1965simplex}, it starts with a set of points that form a simplex -- a generalization of the notion of a triangle or tetrahedron to arbitrary dimensions. On each iteration, the objective function values at the corner points of the simplex determine the worst corner point. The algorithm attempts to replace the worst point by introducing a new vertex in a way that results in a new simplex. Candidate replacement points are obtained by transforming the worst vertex through a number of operations around the centroid of the current simplex: reflection, expansion, inside, and outside contractions.

  \item[-] \textbf{COBYLA} \citep{powell1994direct}: was developed to solve non-linearly constrained optimization problems. This algorithm follows an approach similar to the DFO method \citep{conn1996algorithm,conn1997recent}, but it uses a linear model approximation for the objective function and constraints, interpolating at the vertices that form a simplex and where a trust-region bound restricts the variables perturbation. Thus, a new vector of variables is calculated which may replace one of the current vertices, either to improve the shape of the simplex or because it is the best vector that has been found so far according to a merit function that gives attention to the greatest constraint violation. The trust-region radius is never increased, and it is reduced when the approximations of a well-conditioned simplex fail to yield improvement to the variables until the radius reaches a prescribed tolerance value that controls the final accuracy.
  
  \item[-] \textbf{NEWUOA}: is an unconstrained optimization method using a quadratic interpolation approximation. Like the DFO method, it seeks to calculate the least value of an objective function by applying the trust-region iteration for adjusting the variables. Now, as we mentioned above, all $n$-dimensional quadratic models have $\left(n+1\right)\left(n+2\right)/2$ parameters. This means that, unless other conditions are imposed,  we require this number of interpolation points to build them. However, in NEWUOA this is an input parameter denoted by $m$. In \cite{powell2006newuoa}, the author proposed to use a quadratic model relying on fewer than $(n + 1)(n + 2)/2$ interpolation points. The remaining degrees of freedom in the interpolation are determined by minimizing the change to the Hessian of the surrogate model between two consecutive iterations. The latter is an advantage since a DFO algorithm aims to use fewer evaluations of the objective function.  
  
  \item[-] \textbf{BOBYQA}: is an iterative algorithm for finding a minimum of an $n-dimensional$ function subject to box-constraint. BOBYQA is a extension of NEWUOA, based on a quadratic interpolation approximation --- see \cite{powell2009bobyqa}.
  
\end{itemize}

The development of derivative-free algorithms dates back to the works of \cite{spendley1962sequential} and \cite{nelder1965simplex} with their simplex-based algorithms. An excellent review and numerical comparisons of state-of-the-art algorithms can be found in \cite{SimReview}.

\section{Results and Discussion}
\label{S:4}
In this section, we analyze and discuss the main results obtained following the two adjustment schemes described in Section \ref{S:3}. First, we report and compare the performance of Cell2Fire with and without tuning the $\vec{x}$ vector when dealing with our case study instance, but using the fire scars obtained from Prometheus hourly simulations for finding $\vec{x}^{*}$ both in a full information (after a wildfire) and partial information (real-time on-going wildfire) approaches. Subsequently, we will review how the adjustment behaves compared to the real historical fire scar. In this case, we do not have the evolution of the fire, so we only adjust the parameters using the final scar.

Finally, a detailed analysis and discussion of the simulation results obtained for Dogrib's 2002 fire in Canada when applying the described tuning approaches with different weight vectors $\vec{\mu}$ is performed and a benchmark of the DFO algorithms applied is shown.

All experiments have been conducted in a daily-use laptop with a 4th generation I7 CPU (1.9 GHz, 2 cores), 8 GB of RAM, and Ubuntu 14.0 OS. 

\subsection{ROS adjustment factors via Prometheus proxy}
In these experiments, the $RAF$ problem is solved with full information of the wildfire of interest in order to adjust the $\vec{x}$ vector. This way, the full evolution of the wildfire is provided with the aim of adjusting the simulator parameters with historical data. Notice that the optimization framework can be easily extended to include multiple historical fires by modifying the objective function in order to obtain a representative set of optimal parameters $\vec{x}^*$ for a certain region/area, training and evaluating the model performance with both a training and testing set of fires to avoid over-fitting, following a machine learning model scheme. 

\subsubsection{Global approach results: Dogrib-North}

Starting with $\vec{x}_0 = (1,1,1,1)$, we use the fire scars generated with Prometheus simulator using the Dogrib instance with the same weather conditions as in the real fire but using a constant wind direction (North) for visualization purposes, recording the hourly fire scars obtained 7 hours after its ignition (full information). We set $\mu_i = 1/7, i = 1,...,7$ to represent the average error between the observed scars and the ones simulated by Cell2Fire. We solve the $RAF$ problem using the algorithms presented in section \ref{Algos}, focusing the discussion on the results obtained using the BOBYQA algorithm. Later, we will perform a detailed comparison with the other algorithms mentioned in the previous section.

Because Cell2Fire and Prometheus both rely on the FBP System \citep{Pais2019}, the initial error value $\varepsilon_{T=7}(\vec{x}_0, \vec{\mu})$ is very low (approx. 36.86 using the euclidean norm). This represents the average difference in the number of cells after 7 hours of fire evolution. Since the total burned area is 3,777 cells (or hectares), then, the percentage difference is 0.97\%. Subsequently, after 107 evaluations of the objective function, a minimum error value of 29.10 (0.77\%) is reached. In this case, the best algorithm (BOBYQA) converged after 9 minutes.

The optimal values are very close to the starting point $\vec{x}_0$, which was to be expected since both Cell2Fire and Prometheus obtain the ROS values from the FBP System. The best parameters $\vec{x}^*$ are: $HFactor = 1.41$, $FFactor = 1.16$, $BFactor = 1.85$, $EFactor = 1.16$. Similar results in both computational time and accuracy are obtained for multiple wind directions (8 main axes plus the original weather of the case study).

\subsubsection{FMS approach results: Dogrib-North}
\begin{figure}[h!]
	\centering
    \includegraphics[scale=0.4]{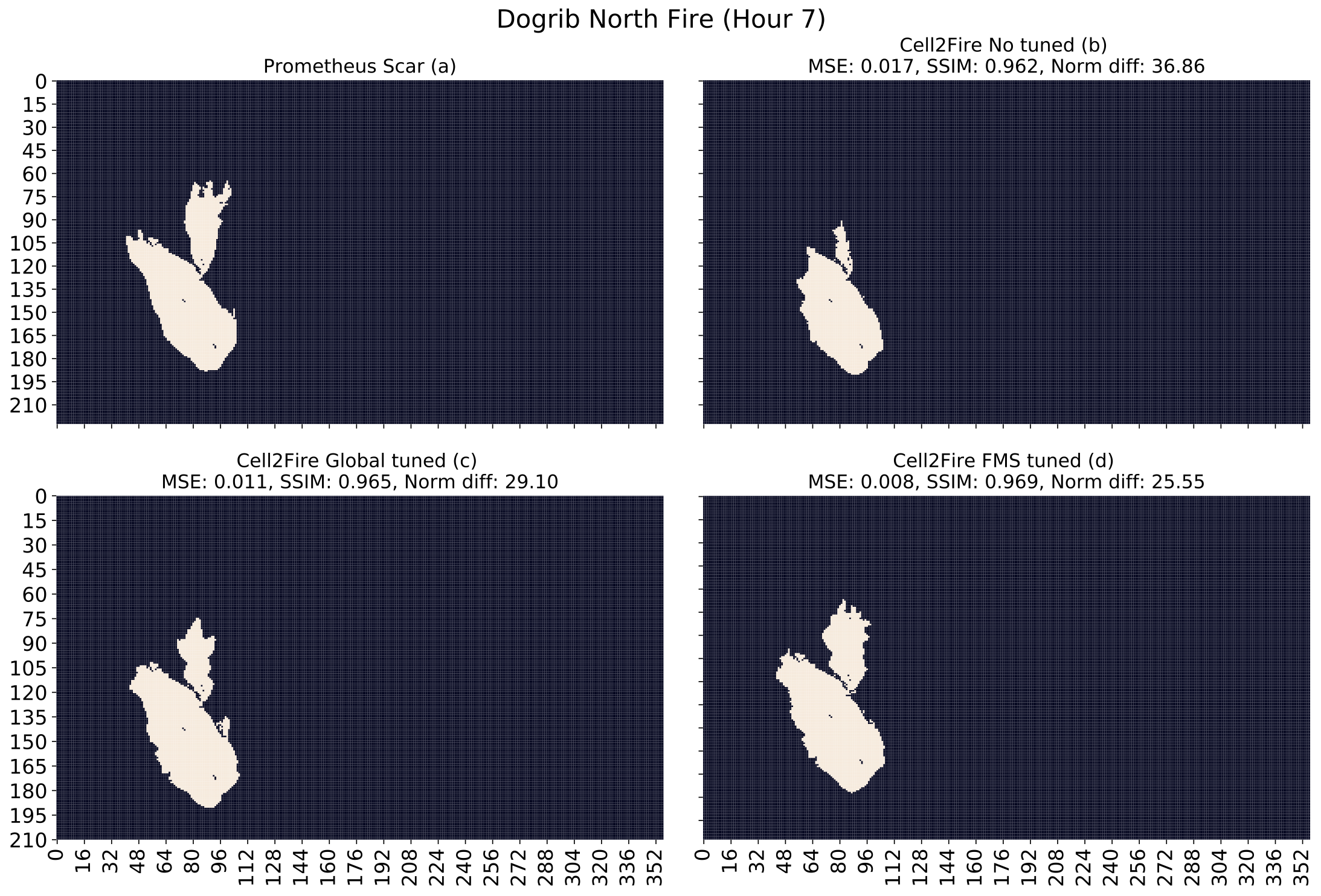}
    \caption{Dogrib fire scar evolution comparison using a constant wind (North) generated by Prometheus (a), Cell2Fire without adjustment (b), Cell2Fire using global tuning (c), and Cell2Fire using a FMS approach (d).}
    \label{Figure5}
\end{figure}

In this section, we address the $RAF_{FMS}$ problem with the Fuel Model Specific (FMS) approach. Following the discussion of section \ref{S:3}, each fuel model $\omega \in \Omega$ of the Dogrib instance corresponds to a ROS factor 4-tuple. This way, since the Dogrib forest contains 8 different fuel types, the $RAF_{FMS}$ problem has 32 variables, in comparison to the 4 variables of the global approach. The optimal factors are shown in Table \ref{Tabla1} and Figure \ref{Figure5} shows the fire scar generated by Prometheus (a), Cell2Fire without adjustment (b), the global (c), and the FMS (d) tuning approaches. Graphically, figures (a), (c), and (d) are very similar, however, without adjustment (b) the fire in the last hour of simulation lacks intensity (strength). This occurs just as the fire crosses the river and is later found in a zone of fuel type C-1. To this respect, important is to mention that Prometheus includes both breaching and spotting phenomena when simulating wildfires while Cell2Fire does not include them in its first version. Therefore, the FMS approach determines that an increase in the magnitude of the C-1 $HFactor$ from 1 to 1.495 is a better approximation (strategy) to decrease the error between both fire scars.  

\begin{table}[h!]
\centering
\begin{adjustbox}{max width=0.88\textwidth}
\begin{tabular}{c|c|c|c|c}
  \textbf{Fuel} & \textbf{HFactor} $\left(x_{1}^{*}\right)$ & \textbf{FFactor} $\left(x_{2}^{*}\right)$ & \textbf{BFactor} $\left(x_{3}^{*}\right)$ & \textbf{EFactor} $\left(x_{4}^{*}\right)$  
  \tabularnewline
  \hline 
    C-1  & 1.495 & 0.786 & 0.995 & 1.424      \\
    C-2  & 1.281  &  1.064 &    1.047 &    1.238   \\
    C-3  & 1.287 & 1.324 & 1.000 & 1.722  \\
    C-4  & 1.002 & 1.000 & 0.998 & 1.002 \\
    C-5  & 0.998 & 1.003 & 1.001 & 1.001   \\
    D-1  & 0.998 &   1.010 &   0.992  &  0.998  \\
    O-1a & 1.219  &  1.055  &  0.995  &  1.125 \\
    M-1  & 1.445 & 1.449 & 1.004 & 1.560 \\
  \hline
  \end{tabular}
  \end{adjustbox}
\caption{Optimized \textit{ROS} adjustment factors (\textit{HROS, FROS, BROS,} and \textit{ECC}) for each fuel type available in the Dogrib instance using Prometheus proxy and BOBYQA algorithm.}
\label{Tabla1}
\end{table}

In contrast, we observe that parameters associated with certain fuels like C-4, C-5, and D-1 do not suffer significant modifications --- they are almost identical to the default parameters $\vec{x} = (1,1,1,1)$. This pattern is due to the fact that Cell2Fire is able to capture similar propagation patterns as the ones modeled by the wave-front propagation approach of Prometheus, as mentioned in \cite{Pais2019}.

Lastly, we observe that the performance in terms of accuracy is superior to the global approach --- as expected --- since each fuel has its own adjustment factors adding more degrees of freedom to the spread model. The minimum value reached using BOBYQA as the main algorithm for this approach is 25.55, i.e. an error of 0.67\%. However, in this case, the number of evaluations of the objective function is 392, which increases the computation time by a factor of 4.

\subsection{Real-Time adjustment}
In this second set of experiments, only partial information is provided when solving the $RAF$ problem. Assuming the presence of an on-going fire, we aim to train and evaluate the performance of the framework in an iterative process where the optimal parameters are updated with new information (fire scars), collected every time-step $t$. This way,  we (1) train the model with the information available up to time $t$ obtaining $\vec{x}_t^*$, (2) simulate the fire scar for $t+1$ using $\vec{x}_t^*$, (3) compare its performance with the observed data at $t+1$, and (4) obtain the $\vec{x}_{t+1}^*$ using $\vec{x}_{t}^*$ as the starting point for the optimization procedure. This process is repeated until a convergence criterion is achieved or no more data is provided.

This setting is significantly relevant in practical applications. As an example, firefighters and decision-makers need to establish their strategy in order to contain an on-going fire. Having an automatic system that learns from the currently available data will improve and support the simulator outputs, becoming a fundamental input for their action plan as more accurate predictions are achieved.

\subsubsection{Global approach results}
As in the previous section, we start from the default parameters $\vec{x}_0 = (1,1,1,1)$, but in this case, we use the 7 hours fire scars generated with Prometheus simulator for the Dogrib North instance to update the optimal parameters every time-step $t$ -- in an iterative process -- using the current optimal parameters as a starting point when solving the $RAF$ problem. This way, we initially update the parameters $\vec{x}_0$ providing the first-hour fire scar $\Pi_1$. Once the optimal vector $\vec{x}^{*}_{1}$ is obtained, we proceed to the next hour ($t = 2$) providing $\Pi_2$ and starting the optimization procedure using $\vec{x}^{*}_{1}$, and so on. Therefore, for adjusting the parameters at time $t$, we use the optimal vector $\vec{x}_{t-1}$ as the starting point when solving the optimization problem, including the last observed scar $\Pi_t$.  

At time $t$, we set $\mu_t = 1/t, i = 1,...,t$ to represent the average error between the observed scars by time $t$ and the ones simulated by Cell2Fire. Again, we focus our analysis using the BOBYQA algorithm. 

\begin{table}[h!]
    \centering
    \begin{tabular}{@{}ccccccc@{}}
    \toprule
    \textbf{Hour} & \textbf{HFactor} & \textbf{FFactor} & \textbf{BFactor} & \textbf{EFactor} & \textbf{Initial Error} & \textbf{Final Error} \\ \midrule
    1             & 0.94             & 1.03             & 0.99             & 0.62             & 8.18                   & 3.87                 \\
    2             & 0.90              & 0.89             & 0.99             & 1.01             & 13.45                  & 7.74                 \\
    3             & 0.99             & 0.99             & 0.13             & 1.12             & 16.03                  & 12.24                \\
    4             & 1.20              & 1.00                & 0.12             & 1.25             & 19.08                 & 13.60                 \\
    5             & 1.20              & 1.00               & 0.12             & 1.26             & 23.62                  & 23.60                 \\
    6             & 1.46             & 1.04             & 0.19             & 1.39             & 28.21                  & 25.90               \\
    7             & 1.66             & 1.02             & 0.35             & 1.49             & 27.69                  & 26.77                \\ \midrule
    \textbf{AVG}  & 1.19             & 1.00             & 0.41             & 1.16             & 19.47                  & 16.25                \\ \bottomrule
    \end{tabular}
    \caption{Global approach results for the real-time adjustment experiment. All factors evolution and a comparison between the initial and final objective value (error) per hour are included. }
    \label{RealTimeGlobal}
\end{table}

From the results presented in Table \ref{RealTimeGlobal}, we observe that the most significant adjustment occurs in the BFactor, in special after the second hour of the ignition time. The perturbation along the larger axis of the ellipse is boosted by the increasing pattern on the HFactor parameter, allowing the fire growth model to reach and surpass the river at the north of the ignition area. This is consistent with the results obtained in the experiments discussed in the previous sections. In addition, we observe how both the initial and final error values reduce their gap (maximum of 52.69\% at hour 1, minimum of 0.08\% at hour 5) when increasing the number of simulated hours, indicating that the performance of the $\vec{x}$ vectors starts to converge. The total run-time needed for solving the whole set of scars was 41.6 [min] -- an average of 6 minutes per time-step tuning -- requiring an average of 74 evaluations of the objective function per $t \in T$ to reach convergence (BOBYQA).

In order to account for potential estimation errors in the input data of the model, several replications with multiple weather scenarios and other potential uncertainty sources (e.g. ignition point) should be performed. This way, confidence intervals could be calculated for each parameter of the optimization problem, allowing the researcher to capture and assess the risk of different potential outcomes.

\subsubsection{FMS approach results}
\begin{figure}[h!]
    \centering
    \includegraphics[scale=0.47]{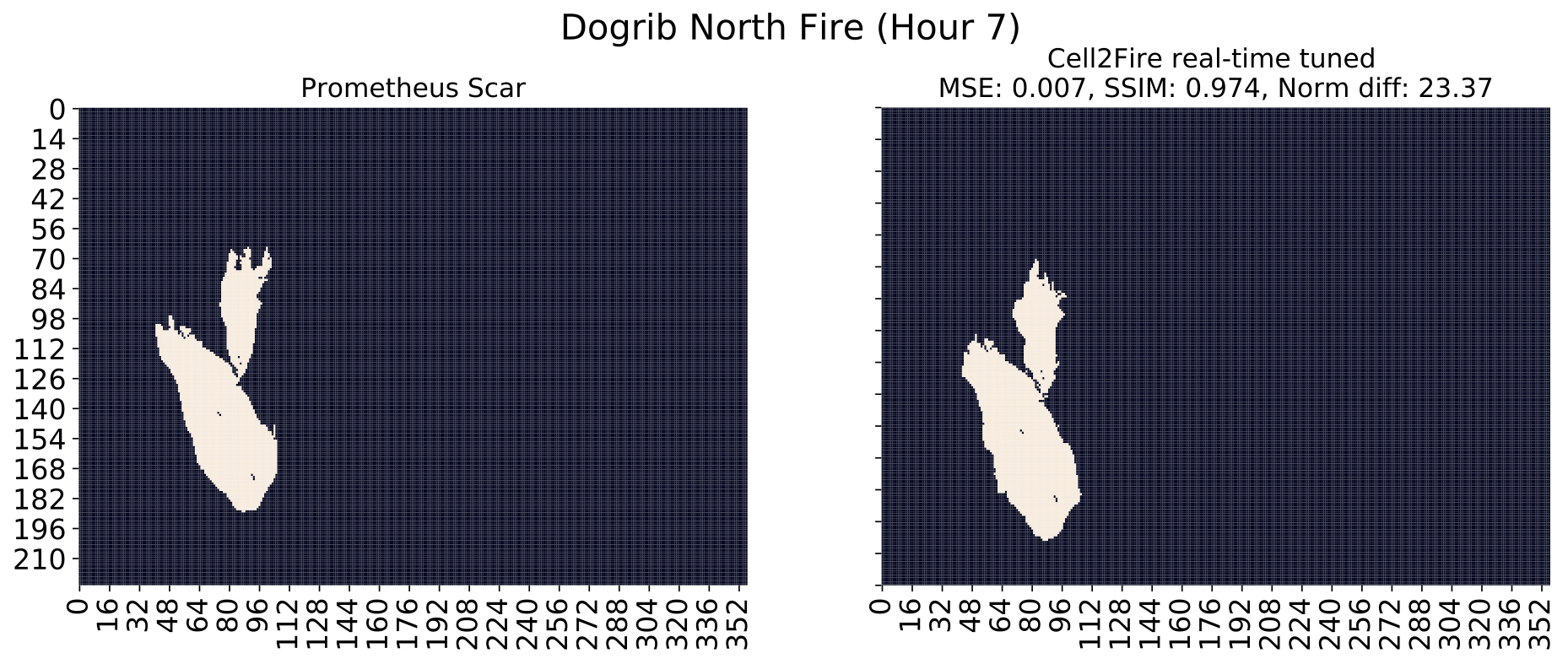}
    \caption{Comparison of Dogrib North's final scars obtained from Prometheus and Cell2Fire real-time FMS adjusted.}
    \label{HourlyFTuned}
\end{figure}

Following the procedure described above, we solve the real-time $RAF_{FMS}$ problem, updating the optimal values of the previous time step vector $\vec{x}^{*}_{t-1}$ once a new scar is available. The average and standard deviation of the optimal factors obtained during the simulation of the 7 hours are shown in Table \ref{TablaHour2} and both Prometheus and Cell2Fire final scars can be seen in Figure \ref{HourlyFTuned}, showing the high-precision of the adjustment method. Graphically, both figures are very similar, reaching an objective value of 23.37, a Mean Squared Error (MSE) of 0.7\% and a very high Structural Similarity  Index (SSIM) of 97.4\%. 

Looking at Table \ref{TablaHour2}, we observe again that the most significant changes occur on C-1 (AVG $HFactor = 1.77 \pm 0.87$, AVG $FFactor = 1.56 \pm 0.55$) and O1-a (AVG $HFactor = 0.69 \pm 0.47$) fuel types, being consistent with the results of the previous sections and showing how the model is able to balance the effect of a higher HFactor on one fuel by decreasing it on another (C-1 vs O1-a). In addition, an average running time of 17 minutes and 340 evaluations per hour simulated are needed by BOBYQA to reach convergence.

\begin{table}[h!]
    \centering
    \begin{tabular}{@{}ccccccccc@{}}
    \toprule
    \multirow{2}{*}{\textbf{Fuel Type}} & \multicolumn{2}{c}{\textbf{HFactor}}                                & \multicolumn{2}{c}{\textbf{FFactor}}                                & \multicolumn{2}{c}{\textbf{BFactor}}                                & \multicolumn{2}{c}{\textbf{EFactor}}                                \\ \cmidrule(l){2-9} 
                                   & \multicolumn{1}{c}{\textbf{AVG}} & \multicolumn{1}{c}{\textbf{STD}} & \multicolumn{1}{c}{\textbf{AVG}} & \multicolumn{1}{c}{\textbf{STD}} & \multicolumn{1}{c}{\textbf{AVG}} & \multicolumn{1}{c}{\textbf{STD}} & \multicolumn{1}{c}{\textbf{AVG}} & \multicolumn{1}{c}{\textbf{STD}} \\ \midrule
    C-1                            & 1.77                             & 0.87                             & 1.56                             & 0.55                             & 1.05                             & 0.02                             & 0.90                             & 0.16                             \\
    C-2                            & 0.98                             & 0.35                             & 0.81                             & 0.06                             & 1.21                             & 0.24                             & 1.40                             & 0.28                             \\
    C-3                            & 1.37                             & 0.29                             & 1.15                             & 0.29                             & 0.98                             & 0.02                             & 1.20                             & 0.22                             \\
    C-4                            & 1.00                             & 0.01                             & 1.02                             & 0.02                             & 1.01                             & 0.03                             & 1.01                             & 0.01                             \\
    C-7                            & 1.00                             & 0.01                             & 1.00                             & 0.01                             & 0.98                             & 0.04                             & 1.00                             & 0.01                             \\
    D-1                            & 1.03                             & 0.06                             & 1.00                             & 0.05                             & 1.00                             & 0.01                             & 1.05                             & 0.15                             \\
    O-1a                           & 0.69                             & 0.47                             & 0.99                             & 0.09                             & 0.99                             & 0.01                             & 1.29                             & 0.68                             \\
    M-1        & 1.02                             & 0.04                             & 1.00                             & 0.01                             & 1.00                             & 0.01                             & 1.03                             & 0.06                             \\ \bottomrule
    \end{tabular}
    \caption{Summary results for real-time Dogrib's adjustment. Average and standard deviation for all factors and fuel types obtained from the 7 hours simulation are shown.}
    \label{TablaHour2}
\end{table}

\begin{figure}[h!]
    \centering
    \includegraphics[scale=0.5]{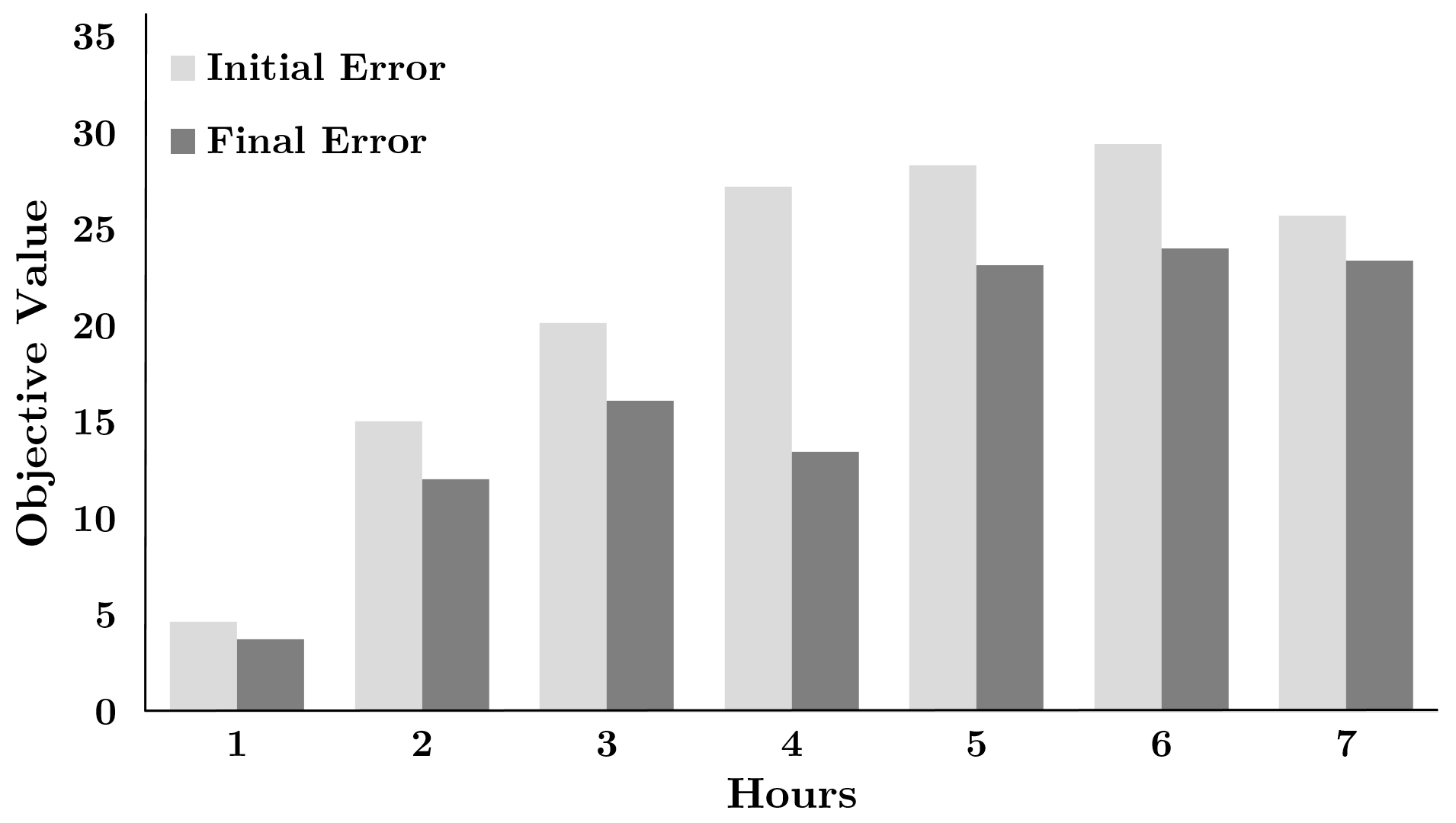}
    \caption{Hourly error evolution using the optimal vector $\vec{x}^{*}_{t-1}$ as the starting point for the time-step $t$ during the real-time tuning FMS approach. Both initial and final errors show a convergence pattern by the end of hour 7.}
    \label{HourlyError}
\end{figure}

Previous results are complemented by Figure \ref{HourlyError}, where the initial and final per-hour errors can be seen. Notice that since the $RAF_{FMS}$ problem has multiple local minima, we can observe situations like in the fourth-hour optimization, where the optimal vector $\vec{x}^{*}_{4}$ reaches a better (lower) objective value than $\vec{x}_{3}^{*}$. In addition, we observe how the method reaches a steady-state in the last 3 hours, being able to stabilize the final error achieved as well as decreasing the initial error towards this value, indicating that the succession of $\vec{x}_t^{*}$ vectors start to converge.

\subsection{Case Study: Dogrib Fire}

The Dogrib fire \citep{Prometheus} started on September 25, 2001,  in the Rocky Mountain front ranges of southwest Alberta. The fire was detected late in the afternoon on September 29 and assessed early the next day at 70 ha. in size. Firefighting started early on October 1. The fire was 828 ha. and out of control on October 15. A wind event resulted in a major fire run on October 16. Local terrain funneled wind flow along the Red Deer River and through a gap in the surrounding mountains. This pushed the fire east along the river valley. The fire jumped the Red Deer River and a road and then resumed a northeast spread direction. The final fire size was 10,216 ha. The October 16 fire run accounted for 90\% of the total area burned and resulted in high to very high burn severities. 

We chose to model this particular fire due to the large documentation and real data availability --- weather conditions recorded from The Yaha Tinda Automatic station, and demographic/topographic data collected from the area --- as well as containing a representative set of different fuel-types documented in the Canadian FBP system. Divided into 79,611 $100 \times 100 \; [m^2]$ cells, we replicate the original fire's ignition point located at (51.652876\degree, -115.477908\degree) starting the fire spreading dynamic on October 16, 2001, 13:00 hrs. For this, the ignition point is translated into an ignition area (cell) in Cell2Fire containing its coordinates.

\subsubsection*{Global approach results}
Starting with $\vec{x}_0 = (1,1,1,1)$  --- default parameters --- we use the real final fire scar from the Dogrib instance obtained 7 hours after its ignition. Therefore, we set ${\left\lbrace \mu_i  \right\rbrace}_{i=1}^{6} = 0$ and $\mu_7 = 1$. We solve the $RAF$ problem using BOBYQA as the main algorithm and compare its performance with COBYLA, NEWUOA, and NELDER-MEAD in section \ref{DFOCOMP}. From the experiments performed, convergence is achieved after 96 evaluations (467.78 [s] = 7.79 [min]) decreasing the objective value from 96.98 to 81.94 (15.5\% improvement), obtaining an MSE = 7.9\% and an SSIM = 0.818. Notice that since we are optimizing a non-convex function, the solution can converge to multiple local optimum values, and thus, it is recommended to perform a series of re-optimizations starting from the best $\vec{\overline{x}}$ (incumbent) obtained or simply starting from different initial points $\vec{x}$ to explore more potential solutions. This is the inherent exploration versus exploitation trade-off, a challenge that is out of the scope of this article and will be covered in future research.

\subsubsection*{FMS approach results}
\begin{figure}[h!]
    \centering
    \includegraphics[scale=0.47]{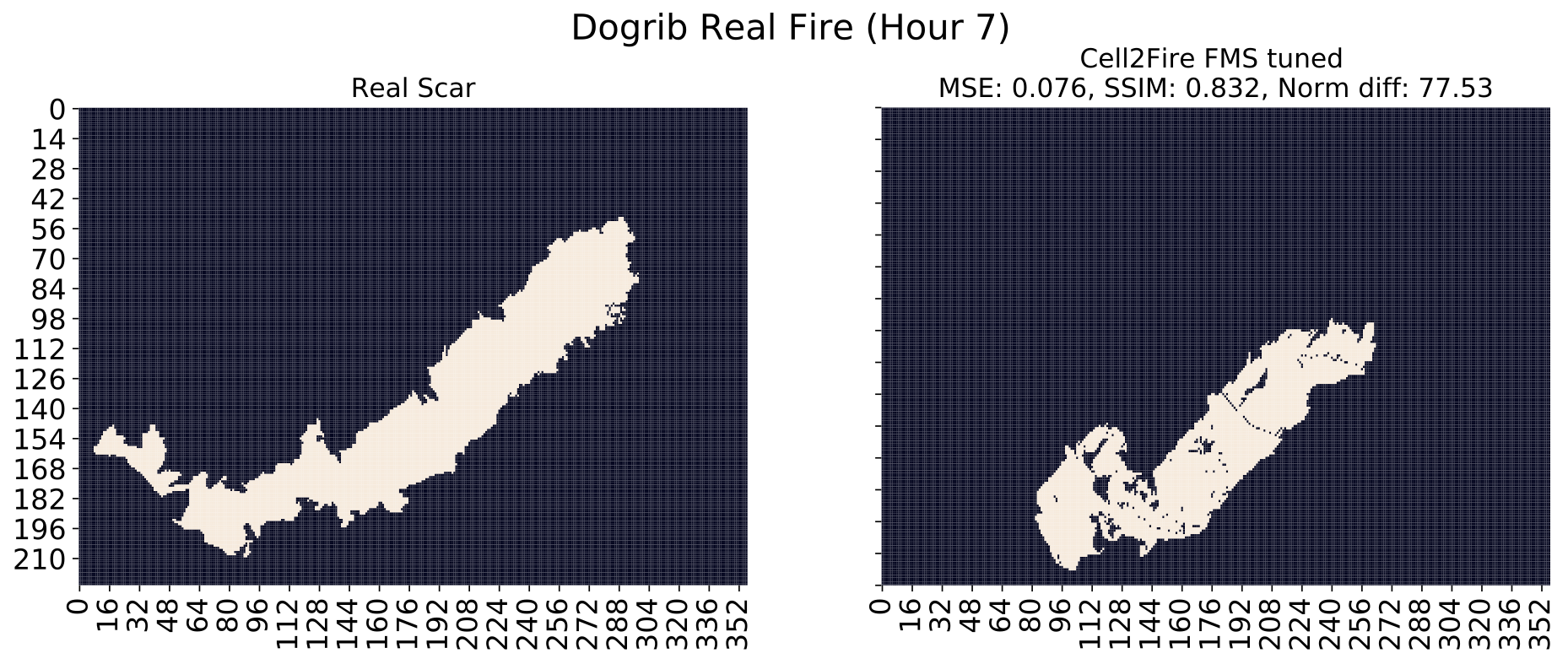}
    \caption{Dogrib real fire scar (left) and Cell2Fire simulated scar using the $\vec{x}^*$ obtained from the $RAF_{FMS}$ problem. A Mean Squared Error value of 0.076 and a Structural Similarity Index value of 0.832 are achieved. The current fire spread and growth model is able to capture the expanding pattern of the real fire in the main direction (north east). However, more detailed information (such as the effect of local wind conditions) is needed in order to capture the back-propagation pattern of the original wildfire.}
    \label{DogribFTuned}
\end{figure}

Below, we show the results obtained from the application of the BOBYQA algorithm in our FMS scheme. Again, we use the starting point $\vec{x_0} = (1,1,1,1)$ and we obtained relevant results in both computation time and final precision with respect to the observed fire. The algorithm converges after 37.58 minutes and the objective function reaches a minimum value equal to 77.53 --- an improvement of 5\% with respect to the global approach. The optimal values obtained for the ROS adjustment factor are shown in Table \ref{Tabla4} and a comparison between the real and simulated scars can be seen in Figure \ref{DogribFTuned}.

\begin{table}[h!]
\centering
\begin{adjustbox}{max width=0.88\textwidth}
\begin{tabular}{c|c|c|c|c}
  \textbf{Fuel} & \textbf{HFactor} $\left(x_{1}^{*}\right)$ & \textbf{FFactor} $\left(x_{2}^{*}\right)$ & \textbf{BFactor} $\left(x_{3}^{*}\right)$ & \textbf{EFactor} $\left(x_{4}^{*}\right)$  
  \tabularnewline
  \hline 
C-1  & 1.623  & 1.243  & 1.023  &  2.479    \\
C-2  & 2.462 & 0.955  & 1.388   & 2.425   \\
C-3  & 2.289 &  0.996 & 1.188  & 1.93  \\
C-4  & 1.158 & 1.138 & 0.287  &  4.91   \\
C-5  & 1.018  & 1.068  &  1.05 & 3.17   \\
D-1  & 0.263  & 2.569  & 0.825  & 1.625  \\
O-1a & 2.066  & 1.675  &  0.269 & 10.82 \\
M-1  & 0.987  & 0.962  & 0.95  &  1.004 \\

  \hline 

  \end{tabular}
  \end{adjustbox}
\caption{Optimized \textit{ROS} adjustment factors (\textit{HROS, FROS, BROS, and ECC)} for each fuel type available in the Dogrib instance from the Canadian FBP system. The tuning has been performed using the real fire scar of the Dogrib instance.}
\label{Tabla4}
\end{table}

Analyzing the results in Table \ref{Tabla4}, it is interesting to notice that certain fuels are significantly affected by the optimal adjustable parameters of $\vec{x}^*$. For example, both the HFactor and EFactor of the O-1a fuel type are significantly modified (1 to 2.066 and 1 to 10.82, respectively) in order to capture its spread pattern based on the real fire scar. Similarly, C-1, C-2, and C-3 fuels are also impacted, mainly in their HFactor, indicating that the default parameters obtained from the FBP system --- given the instance data provided --- are not able to capture the strength of the fire in the main wind direction, and thus, the optimal vector $\vec{x}^*$ tries to adjust the shape of the ellipses, expanding them into the north-east direction.

From the results, we are able to identify significant differences --- as expected --- when using Prometheus or real fire scars. As mentioned, both Prometheus and Cell2Fire use the FBP System as their main fire spread model to determine the relevant ROS values, and differences between both approaches are mainly associated with their fire growth models logics: wave-propagation and cellular automata. However, the initial error with respect to a real fire scar is significantly higher and requires a deeper level of adjustment (more computational time) to find sets of parameters $\vec{x}$ that are able to better capture the real fire propagation dynamics. However, we also remember the fact that historical scars tend to include the effect of suppression activities that are not modeled or included in any of the mentioned simulators, and thus, it plays against the accuracy of the propagation model. In order to account for this bias, suppression efforts should be explicitly modeled by (1) including it as part of the simulation tool or (2) by modifying the data structure of the instance to reflect the position/effect of different activities such as the use of retardant or the location of firewalls, among others.

\subsection{Benchmarking DFO Algorithms}
\label{DFOCOMP}
In this section, we show and discuss the performance of the implemented DFO algorithms, namely BOBYQA, NEWUOA, NELDER-MEAD, and COBYLA. These algorithms have been widely used in different applications -- a review can be found in \cite{SimReview} and \cite{rios2013derivative}. However, to our knowledge, few comparisons in real problems have been reported and even less in ROS adjustment factors in the field of forest fire simulation. 

Our methodology consists of solving the \textit{RAF} problem for the two approaches, Global and FMS, and record the number of evaluations ($NEVAL$) of the objective function until a specified tolerance (error) is reached, which we will denote by $xtol\_abs$. Then, $xtol\_abs$ is a stopping criterion of the algorithm. $\Delta x$ is usually a measure of how much $x$ changes by from one iteration to the next, or the diameter of a search region. This way, the algorithm stops when $|\Delta x| < xtol\_abs$ or a maximum running time (1 hour Global, 3 hours FMS) is reached. This last notation comes from the NLOPT package -- see \citep{johnson2014nlopt} -- a free/open-source library for nonlinear optimization, providing a common interface for a number of different free optimization routines available online as well as original implementations of various other algorithms, in particular: BOBYQA, NEWUOA, COBYLA, and NELDER-MEAD. 

As pointed out in \cite{conn2009introduction,audet2017derivative}, in the Derivative-Free Optimization context, CPU time is irrelevant since we have the number of evaluations as a perfect machine-independent criterion. However, with the propose of supporting decision making in fire fighting, which is usually carried out in real-time, we want this methodology to respond to us as quickly as possible. Therefore, we use two performance indicators: i) the number of Black-Box evaluations ($NEVAL$) and ii) run-time ($RUNTIME$) measured in minutes [min]. In all cases, we execute the algorithms taken as starting point $\vec{x_0} = 1_{\mathbb{R}^{4}}$ in the Global approach or $\vec{x_0}= 1_{\mathbb{R}^{4\left|\Omega\right|}}$ in the FMS approach with $xtol\_abs = 1e^{-16}$.

\subsubsection*{Dogrib-North results}
In Table \ref{Tabla3}, we observe the performance results of each DFO algorithm selected in our study for the Dogrib-North instance, using their default configuration -- internal parameters -- provided by the NLOPT package and our Global tuning approach. In general, the algorithms that stand out most are BOBYQA and NEWUOA in both $NEVAL$ and $RUNTIME$ metrics. Although the one with the best/lowest objective value is NELDER-MEAD, it is the algorithm with the largest run-time (more than one hour) and number of evaluations before achieving convergence, being a critical bottleneck for situations where fast and high-quality responses are required.

\begin{table}[h!]
\centering
\begin{adjustbox}{max width=0.88\textwidth}
\begin{tabular}{c|c|c|c}
   \textbf{Algorithm} & $NEVAL$ & $RUNTIME$ [min]  & $MinValue$   
  \tabularnewline
  \hline 
BOBYQA & \textbf{107} &	18.03 &	29.10 \\
NEWUOA & 125 &	\textbf{16.33} &	26.41\\
NELDER-MEAD & 343 &	67.60 &	\textbf{25.59} \\
COBYLA & 224 &	40.60 &	29.13\\
  \hline 
  \end{tabular}
  \end{adjustbox}
\caption{A summary of the performance of each algorithm is presented in this table for Dogrib-North instance using Prometheus proxy. Best (minimum) results are highlighted.}
\label{Tabla3}
\end{table}

From the results, the NEWUOA algorithm arises as the most suitable method for this particular instance. Several experiments with multiple randomly generated weather conditions were performed, obtaining similar results to the ones already indicated. However, in practice, further improvements can be found by testing multiple configurations of the algorithms, modifying parameters like their tolerance, convergence rate, bounds of the constraints/trust region, among others. This experimentation remains open as part of future research projects.

\subsubsection*{Dogrib real fire results}

Similarly, we show in Table \ref{Tabla7} the performance results of each DFO algorithm selected in our study for the Dogrib real fire, using the Global approach. Again, the algorithms that stand out are both BOBYQA and NEWUOA in terms of the number of evaluations and the total run-time needed to achieve convergence towards a better objective function. As in the previous section, NELDER-MEAD is the algorithm that achieves the best objective value in the instance, mainly due to its default configuration in the NLOPT package, including a larger exploration phase than exploitation, in contrast to BOBYQA and NEWUOA which tend to be more focused on exploitation than exploration. However, both the run-time and the number of evaluations needed by NELDER-MEAD are again very high for time-constrained situations.

\begin{table}[h!]
\centering
\begin{adjustbox}{max width=0.88\textwidth}
\begin{tabular}{c|c|c|c}
   \textbf{Algorithm} & $NEVAL$ & $RUNTIME$ [min] & $MinValue$   
  \tabularnewline
  \hline 
BOBYQA  & \textbf{96} &	\textbf{7.78} &	81.94   \\
NEWUOA    & 104 & 12.23 & 85.56 \\
NELDER-MEAD & 332 &	45.33 &	\textbf{80.54} \\
COBYLA & 217 &	25.56 &	86.58 \\
  \hline 
  \end{tabular}
  \end{adjustbox}
\caption{A summary of the performance of each algorithm is presented in this table. Best (minimum) results are highlighted.}
\label{Tabla7}
\end{table}

Note that the final errors reached by the algorithms -- the minimum value of the $RAF$ objective function -- are similar in magnitude for both instances. However, the error in the real Dogrib fire instance is quite higher than in Dogrib-North. As we already mentioned, this is explained because the fire evolution data is obtained with Prometheus which simulates a fire in its natural form, unlike Dogrib fire's scar which includes the effects of suppression.


\section{Conclusions and Future Work}
\label{S:5}
The adjustment of parameters is a transversal problem in all the disciplines that usually involve the development of software or computer code. In this paper, we have presented a methodology for adjusting the parameters of fire spread models, critical for the accuracy of fire simulators using Derivative-Free Optimization methods that respond to this problem in an efficient way. The DFO-algorithms allowed us to find optimal parameters in the sense that they minimize the evolution error of the hourly fire scars obtained by Cell2Fire and Prometheus. We note at this point that, if real data is available, it would not be necessary to use simulated scars because historical fire data could give us a better fit to the underlying characteristics of the forest under study. However, this is not entirely clear and may produce misleading results because a real scar could be influenced by firefighting actions and thus, the adjusted parameters would contain this latter information (bias) and not just the natural evolution of the fire. In this case, we do not recommend using the real scar for tuning purposes, unless the researchers explicitly include the suppression actions effects, either via extra parameters in the simulation model or by modifying the instance data structure (e.g. introducing new fuel types to model the effect of these activities).

As mentioned, Cell2Fire is a cell-based fire spread simulator that models the fire spread phenomenon via an adaptation of the elliptical model proposed in \cite{ORegan1973,ORegan1976} using the main four ROS components of the FPB System. Nowadays, Prometheus, FARSITE and other simulators widely used are wave-propagation based, and thus, the fire dynamic is modeled in a different approach. Therefore, this methodology can serve as a bridge between the wave propagation approach and the proposed cellular approach in \cite{Pais2019}.  


The proposed tuning framework is able to capture both the local and global factors that affect the fire spread dynamics in a heterogeneous landscape: (1) local effects due to discontinuities of fuel types in the forest are translated to the cell-based model by re-scaling and modifying the shape of the individual ellipses generated for each fuel type and (2) the general structure (spatial disposition, topography, etc.) of the forest is captured by a general re-scaling vector, modifying the strength and magnitude of the fire evolution depending on the characteristics of the heterogeneous landscape. It worth mention that the proposed framework is general enough such that it could be easily adapted to similar simulation tools in different contexts.

Finally, it is important to note that though we have formulated the $(RAF)$ problem using four parameters relative to the rate of spread, we could have used others. In future works, we will focus our research in this direction. It is possible that some parameters such as the slope effect, curing degree, and resilience time are not well adjusted and therefore we could include them as decision variables in the $(RAF)$ problem as well as include new fuels, among other possibilities, in order to improve and extend Cell2Fire to new applications. As a future step, the project will address the development and adaptation of Fuel Prediction Behavior systems for the U.S., Spanish, and Chilean forests based on national field data (work in progress).










\bibliography{Bibliography}

\newpage

\end{document}